# DeltaDiff: Reality-Driven Diffusion with Anchor Residuals for Faithful SR

Chao Yang, Yong Fan*, Qichao Zhang, Cheng Lu and Zhijing Yang

Southwest University of Science and Technology, 621000, China
fanyong@swust.edu.cn

**Abstract.** Recently, the transfer application of diffusion models in super-resolution tasks has faced the problem of decreased fidelity. Due to the inherent random sampling characteristics of diffusion models, direct application in super-resolution tasks can result in generated details deviating from the true distribution of high-resolution images. To address this, we propose DeltaDiff, a novel framework that constrains the diffusion process, its essence is to establish a deterministic mapping path between HR and LR, rather than the random noise disturbance process of traditional diffusion models. Theoretical analysis demonstrates a 25% reduction in diffusion entropy in the residual space compared to pixel-space diffusion, effectively suppressing irrelevant noise interference. The experimental results show that our method surpasses state-of-the-art models and generates results with better fidelity. This work establishes a new low-rank constrained paradigm for applying diffusion models to image reconstruction tasks, balancing stochastic generation with structural fidelity. Our code and model are publicly available at https://github.com/continueyang/DeltaDiff .

## 1 Introduction

Image super-resolution (SR) is a fundamental problem in computer vision that aims to recover high-resolution (HR) images from low-resolution (LR) counterparts. SR tasks require not only the generation of visually pleasing images, but also the reconstruction of accurate HR images that correspond to the given LR inputs. However, because of the irreversibility of the degradation process and the complexity and unknown properties of the degradation kernels in real-world scenarios. SR tasks are an ill-posed problem, and an LR image can correspond to many HR images. These issues make it a long-standing and challenging research field in low-level visual tasks[4, 13].

Currently, diffusion models have gained widespread attention in the field of computer vision due to their exceptional image generation capabilities. Consequently, they have been increasingly applied to SR tasks. Current approaches for applying diffusion models to SR can be divided into two main paradigms. The first involves upsampling the LR images by interpolation before feeding them into the diffusion model[10, 9].

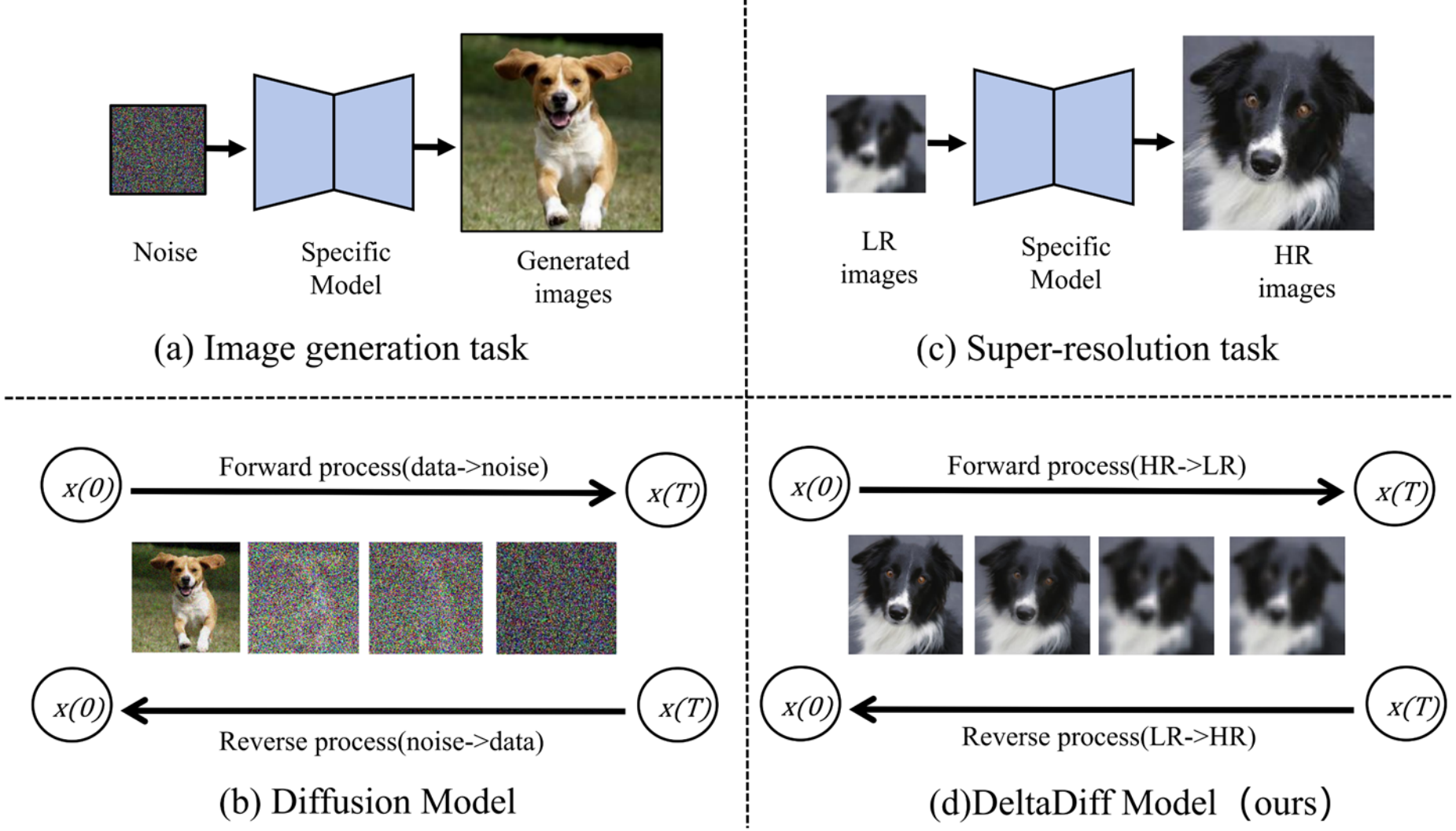

**Fig. 1.**Example figure of the relationships for the model and for the different tasks

During the forward diffusion process, random noise is progressively added to the upsampled images, and in the reverse diffusion process[4, 15] the model removes recover sharp images. The second paradigm uses LR images as conditional inputs to guide the diffusion process.

Notably, existing methods predominantly preserve the original diffusion architecture designed for generative tasks. While the inherent stochasticity in these models enhances output diversity for image synthesis, such characteristics prove suboptimal for SR applications. Specifically, the unmodified diffusion process tends to introduce extraneous high-frequency details that deviate from authentic HR content, thereby compromising reconstruction accuracy.

To address these limitations, recent advancements [9,17,18,12] have proposed integrating LR-HR residual information into diffusion frameworks. These hybrid approaches combine residual guidance with conventional noise prediction while maintaining core diffusion mechanics, demonstrating improved SR performance through targeted architectural modifications. Nevertheless, while introducing residual information, these approaches still retain random noise as the diffusion content and maintain the complete diffusion framework. But, the performance gains from residual guidance have led us to fundamentally re-examine whether noise participation remains essential in the diffusion process.

Examining the diffusion model, its forward process systematically injects random noise into the target image distribution, eventually converging to pure Gaussian noise. The reverse process commences with Gaussian noise and iteratively inverts the forward steps, progressively denoising to reconstruct the original image. While this framework exhibits mathematical rigor particularly effective for image generation tasks, its intrinsic design prioritizes output diversification through stochastic operations, as shown in (a) and (b) of **Fig. 1**. When the diffusion model applied to image generation tasks is applied to super-resolution tasks, the situation changes as super-resolution tasks require a deterministic mapping from LR inputs to their unique HR counterparts. In this mapping, output diversity can have a counterproductive effect on reconstruction fidelity. The qualitative experimental results in our third chapter indicate that the state-of-the-art model using random noise as the expansion path produces detailed information that does not match the real image. To address the fundamental incompatibility with traditional diffusion frameworks, we have redesigned the diffusion mechanism by specifically utilizing image residuals as a transitional medium between the LR and HR domains. We propose DeltaDiff - a residual constrained diffusion model that eliminates random sampling. The diffusion process abandons traditional noise addition and uses HR-LR residual asymptotic replacement as the diffusion mechanism. Its essence is to establish a deterministic mapping path between HR and LR, rather than the random noise disturbance process of traditional diffusion models, as shown in (c) and (d) of **Fig. 1**. Through quantitative and visual comparison experiments with state-of-the-art models, we demonstrate that DeltaDiff enhances the model SR capability while preserving the authenticity of the SR images.

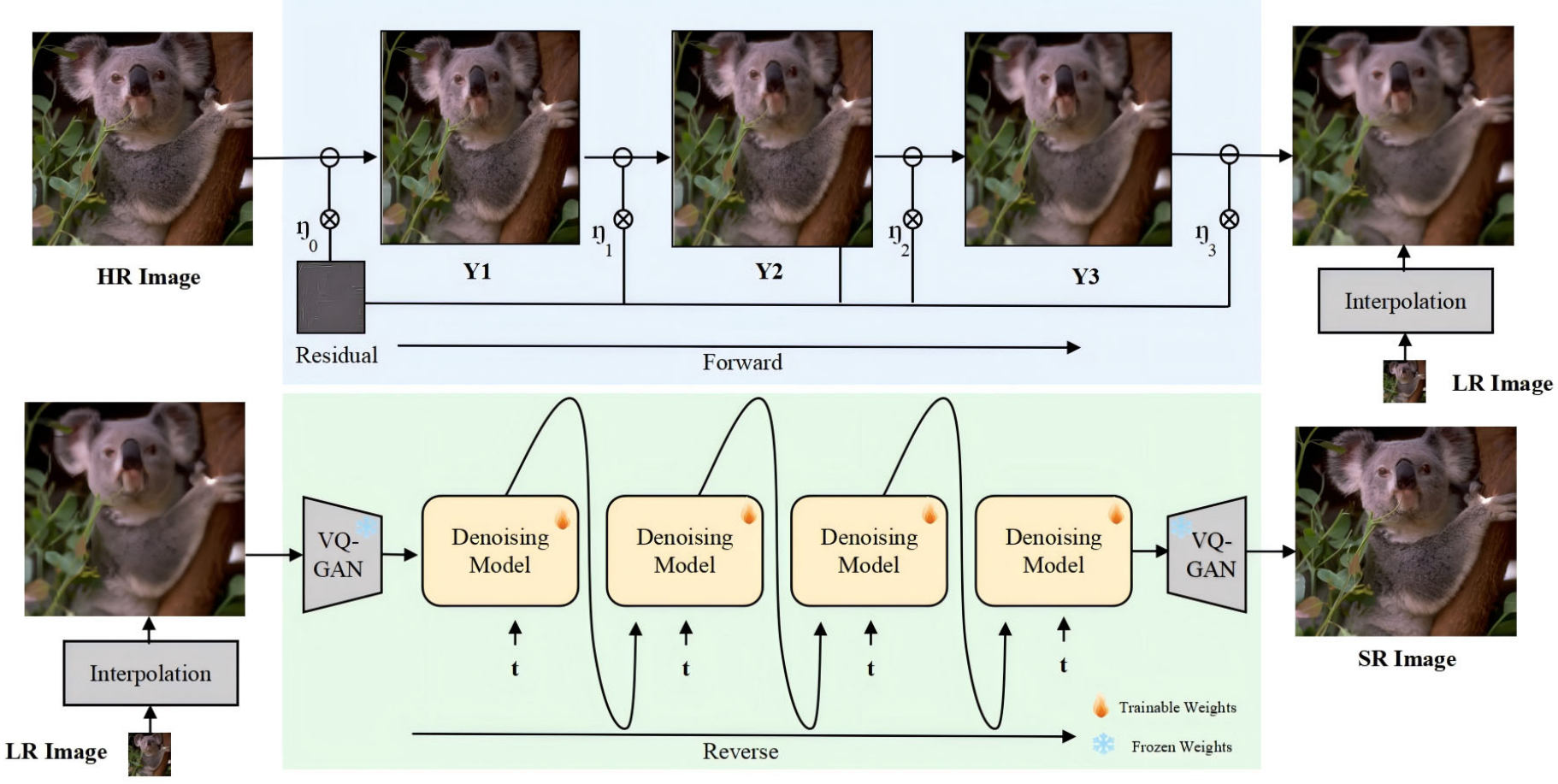

**Fig. 2.** DeltaDiff's deterministic diffusion process: the forward path subtracts controlled residual components from HR to LR, while the reverse process reconstructs HR through residual accumulation.

In summary, the contributions of this paper are as follows.

1)This paper developed a diffusion model for SR, DeltaDiff, whose diffusion process requires only four steps.

2)Establishes a residual-driven diffusion paradigm that eliminates stochastic sampling through direct residual transition modeling between LR and HR domains.

3)Through quantitative and visual experiments, this paper validates the effectiveness of the DeltaDiff model for SR.

# 2 METHODOLOGY

This section presents our proposed DeltaDiff method. This approach involves a diffusion operation based on the residual between the LR and HR images, ensuring that the generated images maintain high fidelity. **Fig. 2** illustrates the overall framework of the method.

## *2.1 Motivation*

The direct application of diffusion models to SR tasks faces a fundamental conflict: the inherent stochasticity of diffusion sampling contradicts the deterministic nature of SR mapping. Traditional diffusion frameworks, designed for generative tasks, progressively inject Gaussian noise into pixel space to learn data distributions. While this mechanism enables diverse image synthesis, it introduces detailed information of HR that deviate in SR applications, as random perturbations corrupt the structural correspondence between LR inputs and their unique HR counterparts.

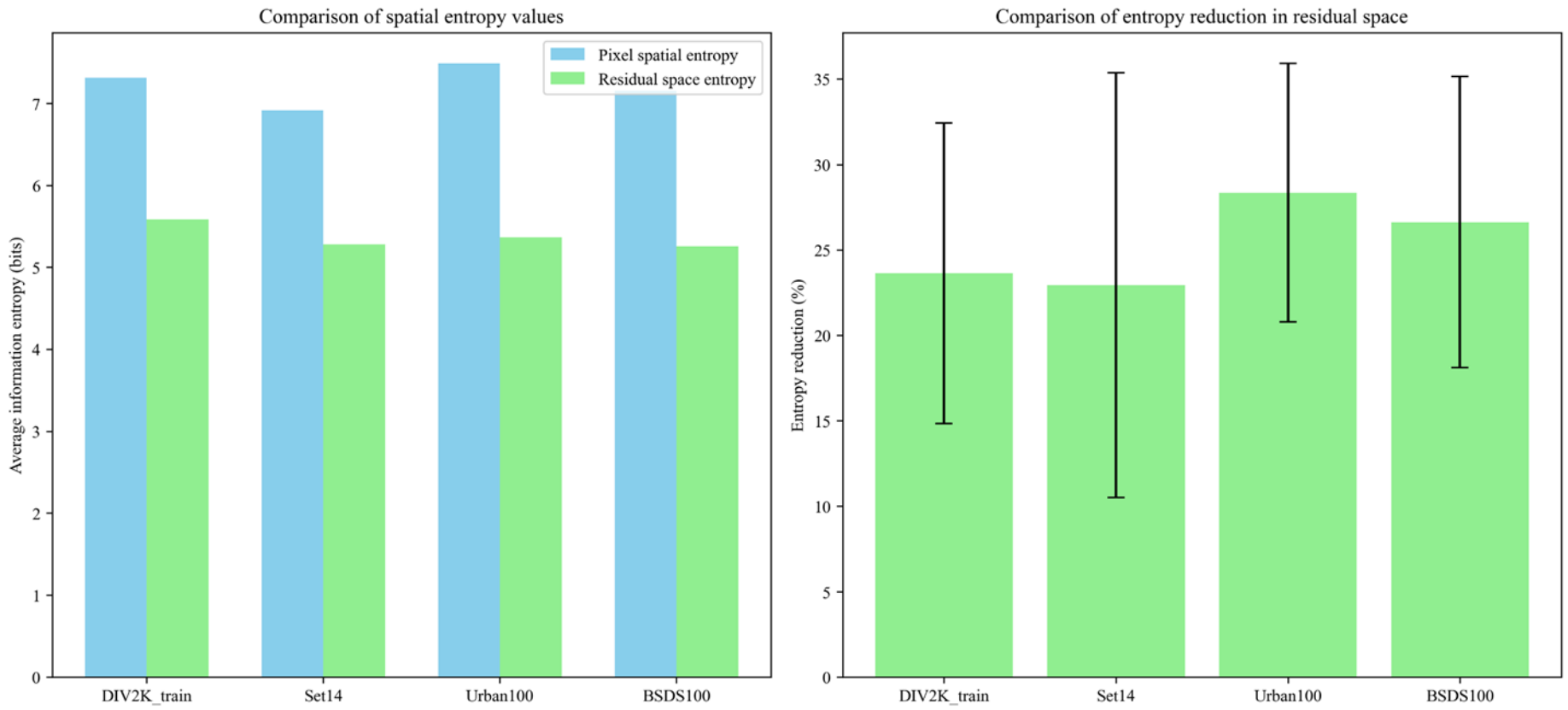

**Fig. 3.** Comparison of entropy reduction percentages in residual space across different datasets, with error bars indicating variability.

To quantify this challenge, we analyze the Shannon entropy of pixel space $S_p$ versus residual space $S_r$. The Shannon entropy of pixel space and residual space can be formalized as:

$$S_p = -\sum_{i=1}^{N} p(x_i) \log_2 p\,(x_i) \tag{1}$$

$$S_r = -\sum_{j=1}^{M} p(r_j) \log_2 p\,(r_j) \quad (2)$$

Where $x_i$ is the pixel intensity and $r_i$ is the residual value. We compare Shannon entropy of $S_p$ and $S_r$ of four standard data sets (DIV2K, Set14, Urban100, BSDS100) used in this paper. Experiments reveal a 24% entropy reduction in residual space demonstrating that residual distributions exhibit significantly lower uncertainty, as shown in **Fig. 3**. This disparity stems from the sparsity and low-rank structure of residuals.

Based on this experimental result, we can conclude that the low entropy residual space has less information and higher certainty, making it easier for the model to learn the deterministic mapping relationship between HR-LR. Therefore, we establish a deterministic mapping path between HR and LR by proposing Deltadiff.

### 2.2 *Model design*

**Forward process.** In order to prevent the model from generating details of illusions, we only use the residual between HR and LR images as the core of our diffusion process to improve image resolution. Unlike previous methods, due to the absence of Gaussian noise in the diffusion process, each step in our forward and reverse processes does not follow a Gaussian distribution. Our diffusion forward process does not target a certain Gaussian distribution, but rather an HR image subtracted by some degree of residual. Specifically, given an LR-HR image pair, our forward process begins with the HR image $\mathrm{y_T}$ and ends at the LR image $\boldsymbol{y_0}$. The forward process is defined as follows:

$$\mathbf{y_{t-1}} = \mathbf{y_0} + \eta_t \cdot (\mathbf{y_t} - \mathbf{y_0}) \quad (3)$$

Here, $\eta$ is a time-dependent factor, such that $\eta_0 \to 1$ (a hyperparameter) and $\eta_t \to 0$ . When $\eta_t \to 0$, the output is equivalent to the HR image.

**Reverse process**. The reverse process fully conforms to the SR process of input LR and output HR. Each step of our reverse process targets images containing different residual information in each step of the forward process. Reconstruct partial residuals at each step of the reverse process. When inputting a blurry image, the reverse process can gradually transition to the final clear image. This process without introducing random factors and distributed sampling greatly ensures the authenticity of the SR image. The definition of the entire reverse process is as follows.

$$\mathbf{y_t} = \frac{\eta_{t-1}}{\eta_t} \cdot \mathbf{x_t} + \frac{\alpha_t}{\eta_t} \cdot \mathbf{y_{t-1}} \quad (4)$$

Here, $x_0$ is the input of the LR image during inference. The sequence $\{\eta_t\}_{t=1}^{T}$ is derived from a moving sequence used in Reshift, which we have adopted. This sequence is increasing monotonically, $\alpha_t$ is the difference between the sequence moving at $\eta_t$ and the previous $\eta_{t-1}$, $\alpha_t = \eta_t - \eta_{t-1}$. From these equations, we can derive the final model output $y_0$ as follows:

$$\mathbf{y_0} = \frac{\eta_0}{\eta_1} \cdot \mathbf{x_0} + \frac{\alpha_1}{\eta_1} \cdot \mathbf{y_1} \tag{5}$$

Due to the fact that the input and output sizes of the diffusion model are the same, HR input can result in extremely high computational resource consumption. To solve this problem, our model adopts the commonly used dimensionality reduction method in diffusion models. Map the input to the hidden space of VQGAN[2] using its encoder, then obtain the output, and restore the image using a decoder. Due to the downsampling process during the mapping process, this greatly alleviates the problem of computational resource consumption. It should be noted that the weights of VQGAN are pre-trained and frozen throughout the training process.

### *2.3 Scheduling Strategy*

We utilized a noise scheduling strategy in diffusion models that allows SR in only four diffusion steps. The adopted schedule is as follows:

$$\eta_t = \sqrt{\eta_1} \times b_0^{\alpha_t}, \quad t = 1,2,\ldots,T-1 \tag{6}$$

$$\alpha_t = \left(\frac{t-1}{T-1}\right)^p \times (T-1) \tag{7}$$

$$b_0 = \exp\left(\frac{1}{2(T-1)} log \frac{\eta_T}{\eta_1}\right) \tag{8}$$

This design ensures that the degradation process within our diffusion framework increases monotonically with time step *t*. Meanwhile, the control of the parameter η provides the ability to control the starting and ending points of diffusion expansion, which makes it possible to balance the fidelity realism of the SR results. We have demonstrated this in the ablation experiment.

## 3 Experiments

In this section, we validate the effectiveness of DeltaDiff by designing quantitative and visual experiments compared to current state-of-the-art algorithms. Since we set the model output size to 256 * 256, we uniformly adjusted the training set size to 256 * 256 and retrained the comparison algorithm using the same settings. Adam[5]optimizer was used and the code was implemented in the Pytorch[8] framework. We set the batch size to 64 and the learning rate to 5e-5. All experiments were performed on a single NVIDIA RTX A6000 GPU (48GB). Quantitative performance evaluation was performed using PSNR[11], SSIM[11] and LPIPS[17] metrics (calculated based on the Y channel). For more details on the experimental setup, please refer to our open-source code.

**Table 1.** Quantitative comparison with state-of-the-art methods on benchmark datasets. The top two results are marked in red and blue

| Method | BSDS100 | | | Set14 | | | Urban100 | | |
|---|---|---|---|---|---|---|---|---|---|
| | PSNR | SSIM | LPIPS | PSNR | SSIM | LPIPS | PSNR | SSIM | LPIPS |
| **BSRGAN** | 21.52 | 0.6065 | 0.3130 | 22.32 | 0.5979 | 0.3219 | 20.08 | 0.5529 | 0.3159 |
| **SwinIR** | 21.03 | 0.5953 | 0.3062 | 21.85 | 0.5828 | 0.3146 | 18.92 | 0.5559 | 0.2936 |
| **DAT** | 20.49 | 0.6150 | 0.4028 | 20.95 | 0.6145 | 0.3965 | 20.93 | 0.5745 | 0.3190 |
| **SinSR** | 23.89 | 0.5985 | 0.4299 | 23.11 | 0.6232 | 0.4014 | 21.21 | 0.6105 | 0.4108 |
| **ResShift** | 23.42 | 0.6196 | 0.3000 | 24.02 | 0.6682 | 0.2717 | 22.49 | 0.6669 | 0.2698 |
| **DeltaDiff** | 23.62 | 0.6373 | 0.2952 | 24.22 | 0.6803 | 0.2716 | 22.79 | 0.6752 | 0.2709 |

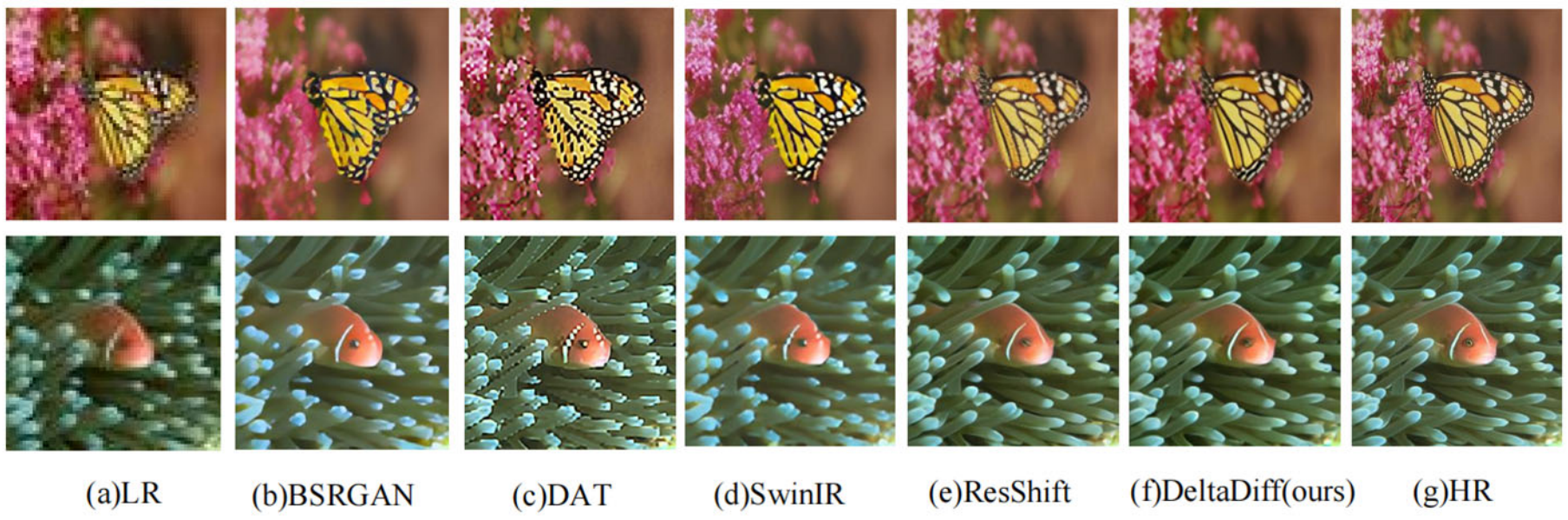

**Fig. 4.** Compared with the state-of-the-art 4x magnification SR visualization algorithm, please zoom in on the image for details.

### 3.1 Ablation Study

We conducted ablation experiments to investigate the impact of two hyperparameters in the diffusion process, $\eta_t$ and $\eta_0$. According to Eq.4, when $\eta_0 \to 1$, the forward process starts with the LR image. When $\eta_t \to 0$ ,the result of the forward diffusion process approaches the HR image. Adjusting these two parameters can change the starting and ending points of the diffusion process. Specifically, during training, when we input an LR image, the input obtained by the model will be LR and $\eta_0$* residual. Increasing $\eta_0$ will increase the residual information obtained in the first step of the diffusion process of the model. And the parameter $\eta_t$ determines the termination target of the final diffusion step of the model. According to Eq.3, when $\eta_t \to 0$, the final result of the forward process is equal to the HR image, and when $\eta_0$<1, the model's result cannot return to the HR image.

To test the effects of different starting and ending points, we performed experiments on the three metrics using four different settings for $\eta_t$ and $\eta_0$, as shown in **Table 2**. The results were optimal when $\eta_t = 0.01$ and $\eta_0 = 0.99$. Adjusting the endpoint $\eta_t$ led to significant image degradation, indicating that the model was unable to restore true high-definition images. Adjusting the starting point $\eta_t$ increased the amount of information received by the model, which reduced the learning difficulty, but resulted in slightly lower generation quality. And when we adjust $\eta_0 = 0.999$, it is closer to the HR image.

Although $\eta_t = 0.5$, the PSNR and SSIM metrics still improved, indicating that the generated images are closer to HR images. But at the same time, the perceptual indicator LPIPS, which represents generalization ability, showed a slight decrease.

We provide visual ablation experiments comparing the forward and reverse diffusion processes between DeltaDiff and previous diffusion processes. The results are shown in **Fig. 5**. From the experimental results, it can easily be observed from the experimental results that the starting point of the forward diffusion process is the HR image, while the end point is Gaussian noise. It is different from the previous diffusion process.

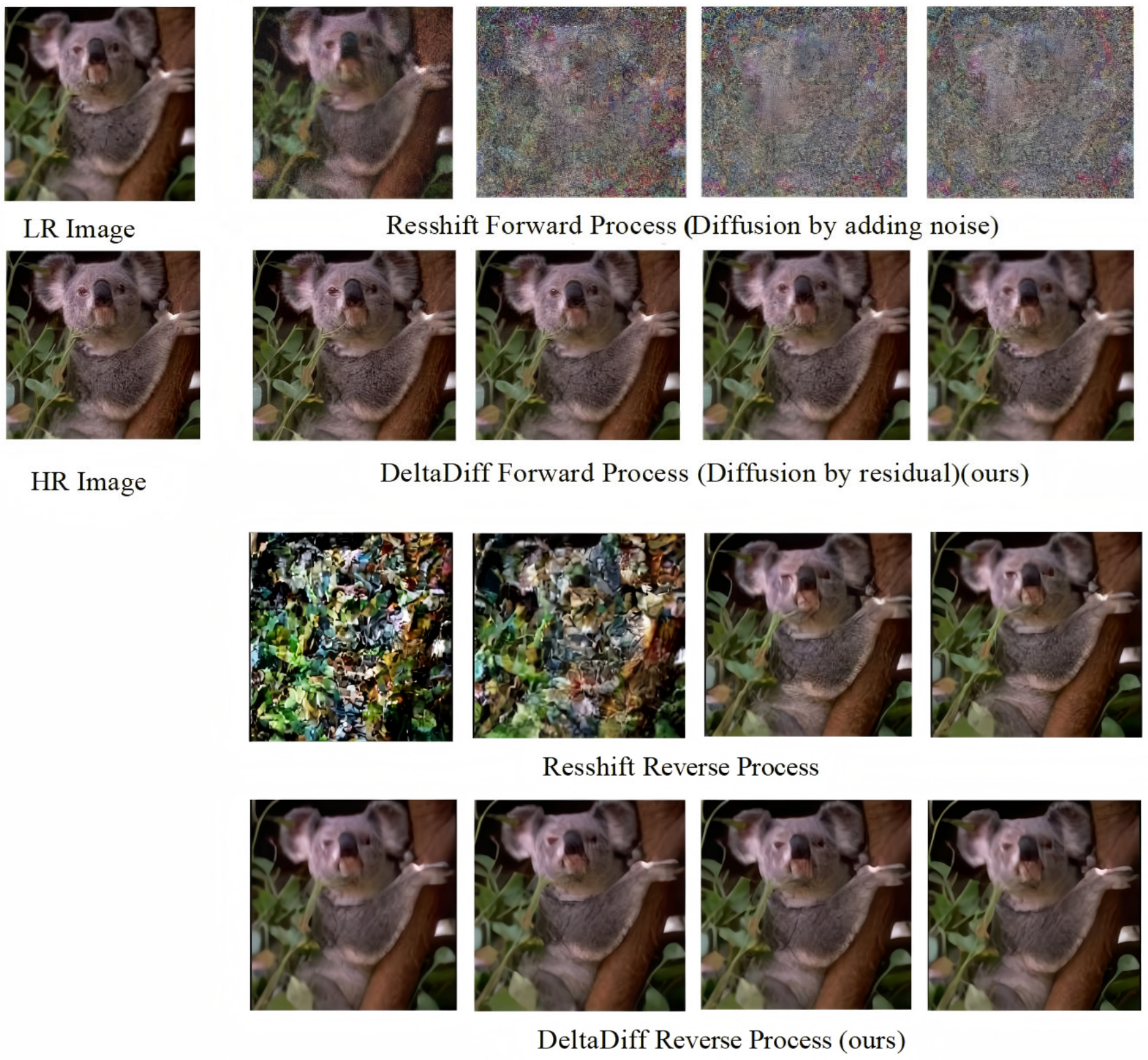

**Fig. 5.** Dynamic comparison visualization of diffusion paths: (a) The forward process of traditional random diffusion methods starts from high-resolution images, gradually degenerates into a noisy intermediate state through multi-step Gaussian noise injection, and finally converges to the LR conditional noise domain. Its reverse reconstruction trajectory exhibits divergent characteristics, with detail shifts occurring in texture dense areas. (b) The residual diffusion path proposed in this article uses the forward process as a LR anchor point and establishes a deterministic mapping through progressive residual compensation. Maintain structural coherence on biological textures and geometric edges.

The starting point of Deltadiff is the HR image, and the ending point is the LR image. Meanwhile, we observe that Resshift diffusion processes, which rely on noise and sampling mechanisms, diffuse images into noisy representations. In the reverse diffusion process, this resembles an image generation process, but it fails to fully utilize the information from the LR images, potentially leading to the generation of fictitious details. DeltaDiff's diffusion process is more stable, allowing for the SR result of more authentic information based on the LR images.

### 3.2 Comparison with State-of-the-Art Methods

To validate the SR capabilities of our method in various types of images, we selected three datasets, Set14[15], BSDS[7], and Urban[3] as our test sets. These datasets were selected to represent varied real-world restoration scenarios. For consistency, we unified the sizes of the test sets to either 256x256 or 512x512. We compared our model with the current state-of-the-art SR algorithms, including BSRGAN[16](CVPR2021), SwinIR[6] (CVPR2021), DAT[1](ICCV2023), SinSR[10] (CVPR2024) and Resshift[14] (NeurIPS2023), with quantitative results reported in **Table 1**. Compared to these advanced algorithms, DeltaDiff achieved the best overall results in all evaluation metrics. Due to the fact that our algorithm did not make any changes to the model architecture, but instead redesigned the entire diffusion process. Therefore, this will not result in an increase in computational complexity or additional memory consumption. However, with over 100M parameters, DeltaDiff still exhibits substantially higher parameter counts compared to traditional non-diffusion SR models (typically <10M parameters), indicating considerable resource requirements.

Notably, during our dataset resizing and model retraining procedures, an intriguing phenomenon emerged: The reduced information complexity in downscaled images created distinct learning patterns. Non-diffusion-based algorithms (BSRGAN, SwinIR, DAT) rapidly reached training plateaus due to diminished learning signals, while diffusion-based counterparts (SinSR, ResShift, DeltaDiff) demonstrated remarkable adaptability. Particularly in low-information scenarios, these diffusion models achieved 10% greater PSNR improvements than conventional approaches through continuous noise modeling, suggesting enhanced capability in capturing subtle details from limited data.

| $\eta_t$ | $\eta_0$ | PSNR | SSIM | LPIPS |
|---|---|---|---|---|
| **0.01** | 0.8 | 23.73 | 0.6379 | 0.3897 |
| **0.01** | 0.99 | 24.58 | 0.6782 | 0.2709 |
| **0.2** | 0.99 | 24.34 | 0.6799 | 0.2784 |
| **0.5** | 0.999 | 24.68 | 0.6950 | 0.2767 |

**Table 2.** Experimental ablation of hyperparameters $\eta_t$ and $\eta_1$ during the forward process. The dataset uses Set5.

To demonstrate the SR capability of our algorithm. We also performed visual comparison experiments on the SR quality of our model against the state-of-the-art algorithms, with the results shown in **Fig. 4**. Compared to non-diffusion-based methods, our approach produced the best SR results. Compared to other diffusion-based models, our method generated results that were closer to the true HR images. Zooming in to examine the details of the images shows that the images generated by our model contain less fictional information than the images generated by the previous diffusion model. For example, the patterns on the butterfly wings and the body details of the fish in our images.

To verify whether our algorithm improves the authenticity of SR results, we compared our algorithm with other algorithms in detail, and the comparison results are provided in **Fig. 6**. By comparing the details, we can find that the previous GAN based network and Transformer based network (such as BSRGAN, SwinIR) lack image details in the SR results. However, algorithms that use diffusion models (such as the Resshift) may have details, but these details do not come from HR images, causing the SR results to lose authenticity. Compared with the above methods, our algorithm has achieved good preservation of details and authenticity. We provide more results of our generated images on our open-source website.

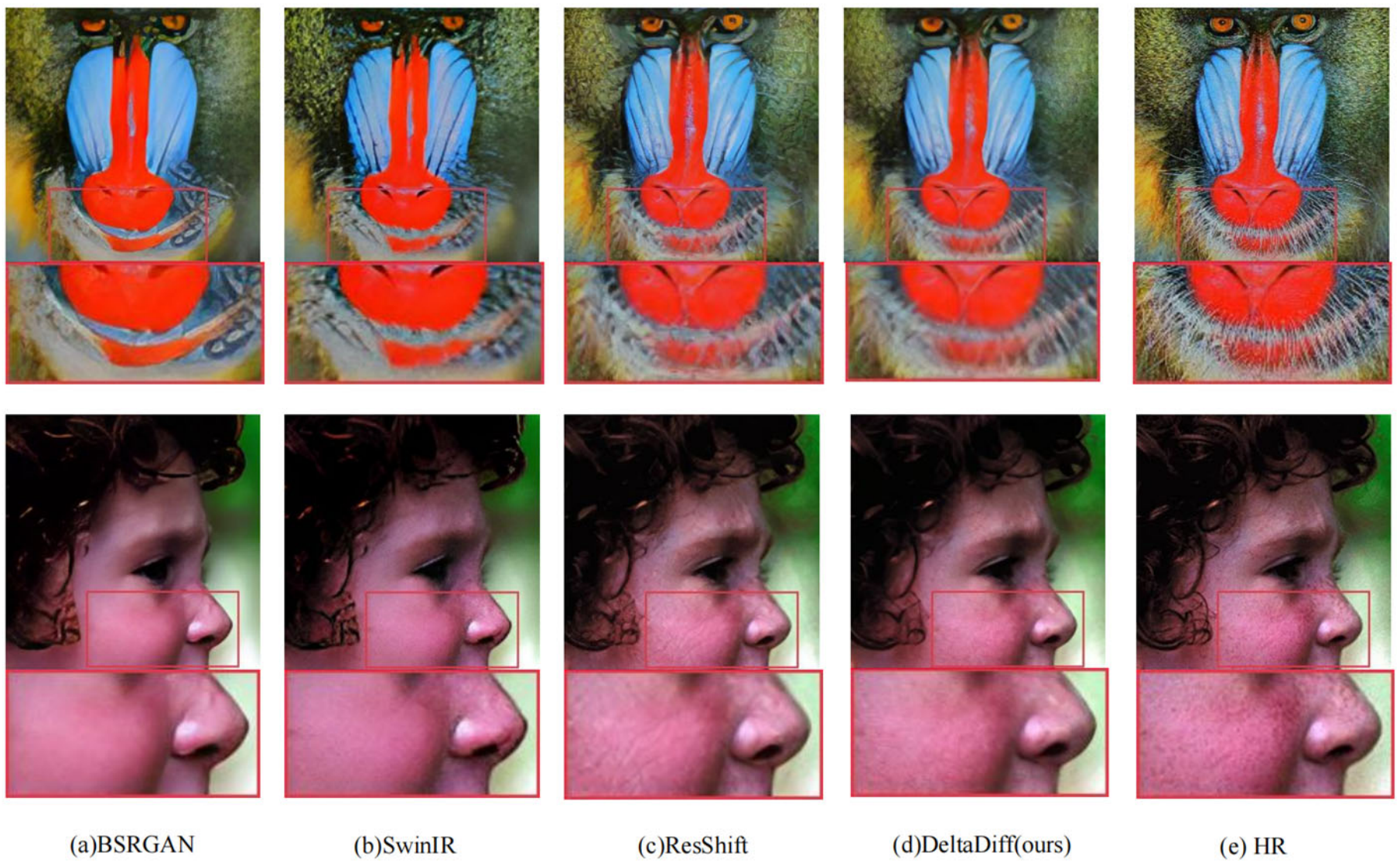

**Fig. 6.** The visualization experiment focuses on the SR capability of details, and zooming in can reveal the degree of difference in details between HR images. The previous diffusion model generates details that do not belong to the HR image, such as (c).

## 4 Conclusion

Existing work that applies diffusion models to SR tasks suffers from retaining random Gaussian noise and sampling mechanisms, resulting in generated images that deviate from true HR information. Our analysis shows that directly applying diffusion models for image generation tasks to the SR domain is suboptimal. We designed a new diffusion process that transitions from LR images to HR images using image residuals. Our diffusion process targets the combination of images and residuals at each step, without introducing noise factors that would increase the randomness of the reconstruction results. Through quantitative and visual comparisons with state-of-the-art models, we demonstrated that our method can effectively enhance reconstruction capabilities while ensuring authenticity.

Our study not only shares the work , but also aims to explore and reflect on a broader question: when combining diffusion models with downstream tasks, is it necessary for the diffusion process to involve Gaussian noise? Our work is currently only focused on diffusion of LR images based on bicubic interpolation. If faced with real images with multiple degradations, there may be some potential shortcomings. Therefore, further work can explore whether different types of degradation can be included in the diffusion process?

## References

[1] Zheng Chen, Yulun Zhang, Jinjin Gu, Linghe Kong, Xiaokang Yang, and Fisher Yu. Dual aggregation transformer for image super-resolution. In *Proceedings of the IEEE/CVF international conference on computer vision*, pages 12312–12321, 2023.

[2] Patrick Esser, Robin Rombach, and Bjorn Ommer. Taming transformers for high-resolution image synthesis. In *Proceedings of the IEEE/CVF conference on computer vision and pattern recognition*, pages 12873–12883, 2021.

[3] Jia-Bin Huang, Abhishek Singh, and Narendra Ahuja. Single image super-resolution from transformed self-exemplars. In *Proceedings of the IEEE conference on computer vision and pattern recognition*, pages 5197–5206, 2015.

[4] Junjun Jiang, Chenyang Wang, Xianming Liu, and Jiayi Ma. Deep learning-based face super-resolution: A survey. *ACM Computing Surveys (CSUR)*, 55(1):1–36, 2021.

[5] Diederik P Kingma. Adam: A method for stochastic optimization. *arXiv preprint arXiv:1412.6980*, 2014.

[6] Jingyun Liang, Jiezhang Cao, Guolei Sun, Kai Zhang, Luc Van Gool, and Radu Timofte. Swinir: Image restoration using swin transformer. In *Proceedings of the IEEE/CVF international conference on computer vision*, pages 1833–1844, 2021.

[7] David Martin, Charless Fowlkes, Doron Tal, and Jitendra Malik. A database of human segmented natural images and its application to evaluating segmentation algorithms and measuring ecological statistics. In *Proceedings Eighth IEEE International Conference on Computer Vision. ICCV 2001*, volume 2, pages 416–423. IEEE, 2001.

[8] Adam Paszke, Sam Gross, Francisco Massa, Adam Lerer, James Bradbury, Gregory Chanan, Trevor Killeen, Zeming Lin, Natalia Gimelshein, Luca Antiga, et al. Pytorch: An imperative style, high-performance deep learning library. *Advances in neural information processing systems*, 32, 2019.

[9] Robin Rombach, Andreas Blattmann, Dominik Lorenz, Patrick Esser, and Björn Ommer. High-resolution image synthesis with latent diffusion models. In *Proceedings of the IEEE/CVF conference on computer vision and pattern recognition*, pages 10684–10695, 2022.

[10] Yufei Wang, Wenhan Yang, Xinyuan Chen, Yaohui Wang, Lanqing Guo, Lap-Pui Chau, Ziwei Liu, Yu Qiao, Alex C Kot, and Bihan Wen. Sinsr: diffusion-based image super-resolution in a single step. In *Proceedings of the IEEE/CVF Conference on Computer Vision and Pattern Recognition*, pages 25796–25805, 2024.

[11] Zhou Wang, Alan C Bovik, Hamid R Sheikh, and Eero P Simoncelli. Image quality assessment: from error visibility to structural similarity. *IEEE transactions on image processing*, 13(4):600–612, 2004.

[12] Bin Xia, Yulun Zhang, Shiyin Wang, Yitong Wang, Xinglong Wu, Yapeng Tian, Wenming Yang, and Luc Van Gool. Diffir: Efficient diffusion model for image restoration. In *Proceedings of the IEEE/CVF International Conference on Computer Vision*, pages 13095–13105, 2023.

[13] Chao Yang, Yong Fan, and Cheng Lu. Dropout multi-head attention for single image super-resolution. In *ICASSP 2024-2024 IEEE International Conference on Acoustics, Speech and Signal Processing (ICASSP)*, pages 2655–2659. IEEE, 2024.

[14] Zongsheng Yue, Jianyi Wang, and Chen Change Loy. Resshift: Efficient diffusion model for image super-resolution by residual shifting. *Advances in Neural Information Processing Systems*, 36, 2024.

[15] Roman Zeyde, Michael Elad, and Matan Protter. On single image scale-up using sparse-representations. In *Curves and Surfaces: 7th International Conference, Avignon, France, June 24-30, 2010, Revised Selected Papers 7*, pages 711–730. Springer, 2012.

[16] Kai Zhang, Jingyun Liang, Luc Van Gool, and Radu Timofte. Designing a practical degradation model for deep blind image super-resolution. In *Proceedings of the IEEE/CVF International Conference on Computer Vision*, pages 4791–4800, 2021.

[17] Richard Zhang, Phillip Isola, Alexei A Efros, Eli Shechtman, and Oliver Wang. The unreasonable effectiveness of deep features as a perceptual metric. In *Proceedings of the IEEE conference on computer vision and pattern recognition*, pages 586–595, 2018.